\documentclass[letterpaper]{article} 
\usepackage{aaai2027}  
\usepackage[hyphens]{url}  
\usepackage{graphicx} 
\usepackage{natbib}  
\usepackage{caption} 
\usepackage{algorithm}
\usepackage{algorithmic}
\usepackage{amsthm}
\usepackage{amsmath}
\usepackage{amsfonts}
\newtheorem{theorem}{Theorem}
\newtheorem{proposition}{Proposition}
\newtheorem{lemma}{Lemma}
\newtheorem{definition}{Definition}
\usepackage{newfloat}
\usepackage{listings}
\DeclareCaptionStyle{ruled}{labelfont=normalfont,labelsep=colon,strut=off} 
\floatstyle{ruled}
\newfloat{listing}{tb}{lst}{}
\floatname{listing}{Listing}

\usepackage{booktabs}

\nocopyright
\newcommand{\lacam}{LaCAM}
\newcommand{\lacamstar}{LaCAM$^\ast$}
\newcommand{\pibt}{PIBT}
\newcommand{\gb}{Goal-Branching}
\usepackage{xcolor}
\definecolor{gbcol}{RGB}{0,90,181}         
\newcommand{\chg}[1]{\textcolor{gbcol}{#1}}
\title{Goal-Branching LaCAM: Temporary Goals as Search Branches in Dense Multi-Agent Pathfinding}
\author {
    Stephen Chang \textsuperscript{\rm 1},
    Daniel Harabor \textsuperscript{\rm 1}
}
\affiliations {
    \textsuperscript{\rm 1}Monash University, Australia\\
    stephen.chang2@monash.edu, daniel.harabor@monash.edu
}

\begin{document}

\maketitle

\begin{abstract}
\lacam{} is a search-based solver for multi-agent pathfinding (MAPF) that runs a depth-first search over joint configurations, using PIBT to generate successors and lazily adding single-agent constraints to explore alternatives. Although \lacam{} is complete, it can struggle on dense, corridor-heavy instances because extracting blocking agents from deadlocks requires coordinated moves of several agents, often away from their goals, which PIBT's greedy one-step successors rarely propose and which \lacam{} can only reach by enumerating constraints one agent at a time. This paper proposes \gb{}, where blocking agents are given temporary goals, each tried as an additional branch of \lacam{}'s search. We examine random and corridor-aware rules for choosing these temporary goals and show that completeness is preserved for any such rule. On dense instances (60--90\% of free cells occupied) of three maze and room maps, \gb{} solves 89\% of instances against 44\% for \lacamstar{}, and, on instances both solve, it finds a first solution 10--14$\times$ faster on maze-32-32-2 and room-32-32-4, and nearly 200$\times$ faster in median time at 75\% density on room-32-32-4. Solution cost increases by 15--22\% on instances both solve, while on open maps and the full MovingAI benchmark its solutions cost no more than those of \lacamstar{} and it is at most 3\% slower.

\end{abstract}

\begin{links}
    \link{Code}{https://github.com/Elysium8/gb-lacam2}
\end{links}

\section{Introduction}
Multi-agent pathfinding (MAPF) asks for collision-free paths that move a team of agents from their start locations to their goals on a shared graph \citep{DBLP:conf/socs/SternSFK0WLA0KB19}. It underpins applications such as automated warehouses, where hundreds of robots share narrow aisles \citep{DBLP:journals/aim/WurmanDM08}, and it is NP-hard to solve optimally \citep{DBLP:conf/aaai/YuL13}. Many of the practically hard instances are dense ones, in which agents occupy a large fraction of the free space and must repeatedly make room for one another.

Optimal and bounded-suboptimal solvers such as CBS \citep{DBLP:journals/ai/SharonSFS15} and EECBS \citep{DBLP:conf/aaai/0001RK21} provide quality guarantees but do not scale to large, dense instances. Scalable solvers trade these guarantees for speed, with different strengths. Rule-based planners such as PIBT \citep{DBLP:journals/ai/OkumuraMDT22} compute one step at a time and handle thousands of agents, but are incomplete. Repair-based methods built on large neighbourhood search, such as MAPF-LNS2 \citep{DBLP:conf/aaai/0001CHSK22} and its extensions \citep{DBLP:conf/aaai/0003D0S25}, iteratively remove collisions from an initial plan and are among the strongest solvers on dense instances, often with better solution quality, but are also incomplete. \lacam{} \citep{okumura2023lacam} takes a third route, a search over joint configurations that uses \pibt{} to generate successors; it is complete and finds first solutions quickly at scale, though these are often costly.

\lacam{} is complete, and \lacamstar{}, its anytime extension, is proven to converge to an optimal solution given sufficient time. However, both can struggle to solve dense instances with dead-end corridors and other bottlenecks in a reasonable time. This is because \pibt{}, the configuration generator underlying both solvers, often deadlocks or livelocks in such settings, and the search is then reduced to a systematic but uninformed enumeration of successors, one agent constraint at a time. Some work has augmented the configuration generator for these situations, such as the swap technique~\citep{okumura2023lacamstar}, which resolves livelocks between pairs of agents by pulling one of them towards a branching vertex where the two can exchange places, and tie-breaking heuristics that discourage agents from blocking each other's routes~\citep{DBLP:conf/socs/OkumuraN25}. However, these extensions only change how agents rank their moves towards fixed goals, and the swap technique handles one pair of agents at a time. Jams that require several agents to move away from their goals at once therefore still fall to the constraint search. In this paper we introduce \gb{}, a novel extension to \lacam{} that resolves such jams directly. When \pibt{}'s step jams, \gb{} uses the agents that \pibt{} failed to move to propose temporary goals for those blocking the way, and tries each choice as a separate branch of the search before \lacam{}'s own constraints. Temporary goals change where agents are heading, not the problem being solved, so the search keeps its guarantees, and the branches are cheap to generate from information \pibt{} already computes.
Our contributions are:
\begin{itemize}
  \item \gb{}, an extension of \lacamstar{} that turns \pibt{} jams into
        temporary-goal branches of the search (Section~\ref{sec:gb});
  \item proofs that \gb{} preserves completeness and eventual optimality, and leaves
        \lacamstar{} unchanged on runs without jams (Section~\ref{sec:theory});
   \item an evaluation showing that on dense maze and room instances \gb{} solves 89\% of instances against 44\% for \lacamstar{} and finds first solutions up to 14$\times$ faster on average, and in some settings over 200$\times$ faster. On open maps and the full MovingAI benchmark it is never more than 3\% slower, its solutions cost no more, and it solves 41 more runs than \lacamstar{} (Section~\ref{sec:exp}).
\end{itemize}

\section{Background}
\subsection{Problem Definition}\label{sec:problem}
An instance consists of an undirected graph $G=(V,E)$ and agents $A=\{1,\dots,n\}$, each with a distinct start $s_i \in V$ and a distinct goal $g_i \in V$. A \emph{configuration} $Q$ places every agent on a distinct vertex; $Q[i]$ is the location of agent $i$, and $Q_s$, $Q_g$ are the start and goal configurations. In one timestep each agent moves to an adjacent vertex or stays. A transition $Q \to Q'$ is valid if no two agents occupy the same vertex ($Q'[i] \neq Q'[j]$) and no two agents swap along an edge ($Q'[i] = Q[j] \land Q'[j] = Q[i]$ is forbidden). A solution is a sequence of valid transitions from $Q_s$ to $Q_g$. We measure its quality by the sum of costs, the sum over agents of the last timestep at which each agent arrives at its goal. $\mathrm{dist}(u,v)$ denotes the shortest-path distance in $G$.

\subsection{\lacamstar{} and \pibt{}}\label{sec:lacam}
\pibt{} \citep{DBLP:journals/ai/OkumuraMDT22} computes one valid transition from $Q$. Agents are processed in a priority order; each takes the unreserved vertex among its neighbours and its current vertex that is closest to its goal, breaking ties at random. If that vertex is occupied by an agent $j$ that has not yet moved, $j$ \emph{inherits} the priority and must move first; if $j$ cannot move, the agent tries its next-best vertex, and fails if none remains. The swap technique of \lacamstar{} additionally lets two agents in a head-on conflict exchange places via a nearby branching vertex.

\lacamstar{} \citep{okumura2023lacam,okumura2023lacamstar} runs a depth-first search over configurations. A node $N$ stores a configuration $N.Q$, its parent and neighbours, costs $N.g$, $N.h$ and $N.f=N.g+N.h$, an agent order $N.\pi$ by decreasing dynamic priority $\mathrm{prio}_i$ (which grows by one each step agent $i$ is away from $g_i$, and resets on arrival), and a constraint tree $N.\mathcal{T}$. Each node of $N.\mathcal{T}$ fixes the moves of the first $k$ agents of $N.\pi$; its root fixes none. Successors of $N$ are generated one at a time by taking the next node of $N.\mathcal{T}$, adding its children (all moves of the next agent in $N.\pi$), and running \pibt{} under its constraints. A new configuration becomes a child node and is pushed onto the stack OPEN; a configuration already in EXPLORED instead has its costs rewired and its node pushed again, or, with small probability, the start node (a random restart). After the first solution, of cost $c^\ast$, the search continues and discards nodes with $N.f \ge c^\ast$.

\subsection{Related Work}
\paragraph{Improving \lacam{}'s configuration generator.} Most work on \lacam{} in dense settings changes how \pibt{} ranks moves. \lacamstar{} adds the swap technique~\citep{okumura2023lacamstar}; LaCAM3 adds Monte-Carlo configuration generation, spread-out guide paths and iterative refinement~\citep{DBLP:conf/atal/000124}; hindrance and regret penalise moves that block a neighbour's route~\citep{DBLP:conf/socs/OkumuraN25}; LG-LaCAM steers \pibt{} along short windowed guide paths~\citep{DBLP:conf/aaai/AritaO26}; and LaGAT learns move preferences with a graph attention network~\citep{DBLP:conf/aaai/JainOAP26}. All of these bias the choice of move towards fixed goals. \gb{} is complementary: it changes what agents are heading for, and offers several alternatives as branches of the search. LaGAT also detects deadlocks, as an agent that has not moved while its neighbours are unchanged, but responds by reverting to \pibt{}'s default preferences rather than by branching.

\paragraph{Temporary goals and evacuation.}
Moving blocking agents aside and restoring them later is a long-standing idea in rule-based MAPF. Push and Swap and Push and Rotate push agents, including ones already at their goals, off a path and return them afterwards~\citep{DBLP:conf/ijcai/LunaB11,DBLP:conf/atal/WildeMW13}, and the Corridor Generating Algorithm evacuates the occupants of a corridor to the nearest free vertex outside it~\citep{DBLP:conf/iros/PertzovskySZ24}. Temporary priorities have been used in \pibt{} to push agents out of dead-end trees~\citep{DBLP:conf/kes/FujitaniYMS22}, and within \lacam{}, BRaP-PIBT gives agents temporary goals in block rearrangement, where goals are anonymous~\citep{DBLP:journals/corr/abs-2509-01022}. These methods either commit to a single rule-based choice or address anonymous goals; none derives temporary goals from detected jams and tries several of them as branches of a complete search.

\gb{} bears a resemblance to the Multi-Agent Corridor Generating Algorithm (MACGA) \citep{DBLP:conf/ijcai/PertzovskiySZF25}, in the sense that both algorithms attempt to resolve corridor blockages by evacuating blocking agents. However, MACGA is a rule-based, polynomial-time planner: it detects corridors from the map's separating vertices, moves their occupants to the nearest free vertices outside the corridor, and commits to a single such plan, which makes it incomplete. \gb{} instead operates inside \lacamstar{}'s search. Evacuations are triggered by jams that PIBT itself detects, rather than by map geometry, and several candidate temporary goals are offered as alternative branches that the search can backtrack over. Because \lacamstar{}'s constraint tree remains as the fallback, completeness is retained. Moreover, MACGA's evacuations cannot pass through agents already following an evacuation plan, and fail in that case, whereas in \gb{} an agent heading to a temporary goal can itself cause further temporary goals to be assigned to the agents blocking its way, so evacuations compose across successive jams.

\section{\gb{}}\label{sec:gb}
At a high level, \gb{} triggers when \lacamstar{}'s PIBT step \emph{jams}, that is, when the unconstrained step returns a configuration that has already been explored. Instead of relying only on the constraint tree to find, one agent at a time, the coordinated moves that resolve the jam, \gb{} gives the blocking agents candidate temporary goals and tries each as a separate branch of the search. Each branch runs one PIBT step under its temporary goals, and the resulting configuration is added as a new successor of the high-level node, tried before the node's constraint tree. Search states, costs, heuristics and the structure of the constraint tree are unchanged, which is why completeness is preserved (Section~\ref{sec:theory}). We first define temporary goals and the extended search nodes, then describe how jams yield branches and how the search uses them, and close with a worked example and the theoretical properties.

\subsection{Temporary and Effective Goals}\label{sec:tg}
\gb{} steers agents through \emph{temporary goals}, which change where an agent is
heading without changing the problem it must solve.
A \emph{temporary goal} for agent $i$ is a record $\tau_i = (v_i, p_i, d_i, a_i)$:
target vertex $v_i \in V$, pusher $p_i \in A$ (the agent whose way $i$ is
clearing), the pusher's distance to its goal when the record is set,
$d_i = \mathrm{dist}(Q[p_i], g_{p_i})$, and age $a_i \in \mathbb{N}$.
We write $\tau_i = \bot$ when agent $i$ holds no temporary goal.
The \emph{effective goal} of $i$ is
\begin{equation}
\hat g_i = \begin{cases} v_i & \text{if } \tau_i \neq \bot,\\ g_i & \text{otherwise.}\end{cases}
\end{equation}
Within a PIBT step, every agent ranks its candidate vertices $u$ by $\mathrm{dist}(u, \hat g_i)$, breaking ties at random, and the swap technique of \lacamstar{} likewise tests effective rather than real goals. An agent without a temporary goal therefore behaves exactly as in \lacamstar{}. Temporary goals affect only which successors PIBT proposes: edge costs and the heuristic $h$ are always computed from the real goals $g_i$, so solution quality and the admissibility of $h$ are measured against the original problem.

Temporary goals are passed from a node to its children and expire on their own. When a successor with configuration $Q'$ is created, every record $\tau_i \neq \bot$ ages by one step, $a_i \leftarrow a_i + 1$, and is released ($\tau_i \leftarrow \bot$) if
\begin{equation}\label{eq:release}
a_i > H \;\lor\; Q'[p_i] = g_{p_i} \;\lor\; \mathrm{dist}(Q'[p_i], g_{p_i}) < d_i - 1 ,
\end{equation}
where $H$ is a hold limit ($H=64$ by default). That is, a temporary goal with a pusher is held until the pusher has reached its goal or has made real progress towards it. Arriving at $v_i$ alone does not release it, so the holder stays out of the way, and $H$ bounds how long any agent can be diverted.
\subsection{Extended Search Nodes}

In addition to the fields of Sec.~\ref{sec:lacam}, \gb{} adds three fields:
\begin{itemize}
  \item $N.\tau = (\tau_1,\dots,\tau_n)$, the temporary goals in force at $N$
        ($\tau_i=\bot$ if agent $i$ has none). A child inherits $N.\tau$ after
        ageing and the release rule (Eq.~\eqref{eq:release}) are applied.
  \item $N.\mathcal{B}$, a queue of goal branches still to be tried at $N$.
  \item $N.\mathit{jam}$, set once a jam at $N$ has been handled, so that
        branches are queued at most once per node.
\end{itemize}
Nodes are still identified by $N.Q$ alone: a configuration reached again is treated as already explored, whatever temporary goals it was reached with, and keeps the $\tau$ of the node that first generated it.

The agent order sorts agents by the key
\begin{equation}
\kappa_i = \mathrm{prio}_i + M\cdot[\tau_i \neq \bot], \qquad M \gg \max_j \mathrm{prio}_j,
\end{equation}
in decreasing order, where $\mathrm{prio}_i$ is \lacamstar{}'s dynamic priority: it grows by one at every step that agent $i$ is away from its real goal $g_i$ and resets once the agent arrives. Holders of a temporary goal are therefore planned first, and constrained first in $N.\mathcal{T}$. Within a goal branch, the agents the branch assigns goals to are placed ahead of all others.

\subsection{Jams and Blocking Pairs}\label{sec:jam}

\begin{definition}[Jam]
The unconstrained PIBT step at node $N$ \emph{jams} if it returns a configuration already in the explored set.
\end{definition}

The unconstrained step is the first one \lacamstar{} takes at $N$, before any constraint from $N.\mathcal{T}$ is applied. A jam is handled at most once per node, recorded by $N.\mathit{jam}$.

When a jam occurs, \gb{} collects candidate blockers from the failures of that PIBT step. Whenever agent $b$ is reached through priority inheritance from agent $p$ (that is, $p$ wants $b$'s vertex) and fails to secure any vertex, the \emph{blocking pair} $(b, p)$ is recorded: $b$ is in $p$'s way and could not move. Failures of agents that were not pushed carry no pusher and are ignored. This yields a list $F$ in the order PIBT recorded the failures. Failed pushes are typical of a jam: in a dense configuration a pushed agent often has nowhere to go, so it and its pusher both stay put, and the step either returns $N.Q$ itself or shuffles back to a configuration explored earlier.

If no push failed, $F$ is empty and is filled by a fallback scan instead: going through the agents in the order $N.\pi$, for each agent $a$ with $N.Q[a] \neq \hat g_a$ we take its neighbour $u$ closest to $\hat g_a$ (ties broken by neighbour order), and if another agent $j$ stands on $u$, the pair $(j, a)$ is added, until $F$ holds $P$ pairs. The scan is rarely needed: on the dense instances of Section~\ref{sec:exp}, a push failed in 97\% of jams with the random rule and 94\% with the corridor rule.

In both cases, only the first $P$ distinct blocking agents in $F$ are kept ($P = 4$ by default). Each of them becomes a candidate for a temporary goal (Sec.~\ref{sec:branches}).

\subsection{Generating Branches}\label{sec:branches}

A \emph{goal branch} $\beta$ is a set of entries $(a, w, p)$, each giving agent $a$ the temporary goal $w$ with pusher $p \in A$. \textsc{MakeBranches} takes the first $P$ distinct blocking agents of $F$ (Sec.~\ref{sec:jam}) in order and, for each such agent $b$ and the pusher $p$ of its first pair in $F$, creates branches using one of two \emph{goal rules}.

\paragraph{Random rule.}
Each of the $K$ branches is $\{(b, w, p)\}$, where $w$ is drawn uniformly from the vertices unoccupied in $N.Q$, or, if a radius $R > 0$ is given, from those within distance $R$ of $Q[b]$. Only the blocking agent's goal changes; the rule uses no knowledge of the map.

\paragraph{Corridor rule.} The corridor rule instead sends $b$ somewhere it is out of $p$'s way. It first works out where $p$ is going: $\pi_p$ is a shortest path from $Q[p]$ to $\hat g_p$, and $T$ is the dead end $p$ must enter, that is, the part of the graph around $\hat g_p$ that $Q[b]$ cuts off from the rest, provided it is a tree. A \emph{pocket} is a vertex $b$ can reach from $Q[b]$ without stepping on $\pi_p$ or entering $T$; a breadth-first search finds the $K$ nearest, preferring free vertices and vertices that are no agent's goal. The rule then queues two kinds of branch, in this order:
\begin{itemize}
  \item \emph{Retreat}, if none of the pockets is free. Since $b$ has nowhere to go, $p$ makes way instead: a search from $Q[p]$ that does not pass $Q[b]$ finds up to $K$ vertices off $b$'s shortest path to its goal, and a branch $\{(p, w, b)\}$ sends $p$ to each, so that $b$ can pass.
  \item \emph{Pockets}. One branch $\{(b, w, p)\}$ per pocket $w$ sends $b$ there. If $w$ is occupied by an agent $c$, the branch also sends $c$ to its own nearest pocket, with $b$ as pusher (\emph{recursion}).
\end{itemize}
Each blocking pair therefore yields at most $K$ branches with the random rule and $2K$ with the corridor rule, so a jam queues at most \begin{equation} |N.\mathcal{B}| \le B_{\max} = \begin{cases} PK & \text{(random)},\\ 2PK & \text{(corridor)}\end{cases} \end{equation} branches. We use $P = 4$, $K = 3$ and $R = 0$ (no radius) by default.

\subsection{Search with Goal Branches} \label{sec:search}
\begin{algorithm}[t]
\caption{\lacamstar{} with \gb{} (changes in \chg{blue})}\label{alg:gb}
\renewcommand{\algorithmicif}{\chg{\textbf{if}}}
\renewcommand{\algorithmicthen}{\chg{\textbf{then}}}
\renewcommand{\algorithmicendif}{\chg{\textbf{end if}}}
\begin{algorithmic}[1]
\STATE OPEN, EXPLORED $\gets \{N_s\}$, with \chg{$N_s.\tau_i = \bot$ for all $i$}
\WHILE{OPEN $\neq \emptyset$}
  \STATE $N \gets$ top(OPEN)
  \STATE \textbf{if} $N.\mathcal{T} = \emptyset$ \OR $N.f \ge c^\ast$ \textbf{then} pop OPEN; \textbf{continue}
  \STATE \textbf{if} $N.Q = Q_g$ \textbf{then} record it, update $c^\ast$; \textbf{continue}
  \IF{\chg{$N.\mathcal{B} \neq \emptyset$}}\label{line:br-start}
    \STATE \chg{$\beta \gets$ pop($N.\mathcal{B}$); $\tau' \gets$ apply($\beta$, $N.\tau$)}
    \STATE \chg{$(Q', \cdot) \gets$ PIBT($N.Q$, $\tau'$, $\beta$'s agents first)}
    \STATE \chg{\textbf{if} $Q' \neq \bot$ \textbf{then} \textsc{Insert}($Q'$, $N$, $\tau'$)}
    \STATE \chg{\textbf{continue}}\label{line:br-end}
  \ENDIF
  \STATE $L \gets$ next($N.\mathcal{T}$); expand $L$
  \STATE $(Q', \chg{F}) \gets$ PIBT($N.Q$, \chg{$N.\tau$}, constraints of $L$)
  \IF{\chg{$L$ is the root \textbf{and} the step jams \textbf{and} $\neg N.\mathit{jam}$}}\label{line:jam}
    \STATE \chg{$N.\mathit{jam} \gets$ true; $N.\mathcal{B} \gets$ \textsc{MakeBranches}($N$, $F$)}
  \ENDIF
  \STATE \textbf{if} $Q' \neq \bot$ \textbf{then} \textsc{Insert}($Q'$, $N$, \chg{$N.\tau$})\label{line:insert}
  \STATE \chg{\textbf{if} the step jammed, $Q' \notin \{\bot, N.Q\}$ and $N.\mathcal{B} \neq \emptyset$ \textbf{then} push $N$}\label{line:push}
\ENDWHILE
\end{algorithmic}
\end{algorithm}

Algorithm~\ref{alg:gb} shows the search; text in blue marks \gb{}'s additions and changes to \lacamstar{}. The search is a depth-first search over configurations: $c^\ast$ is the cost of the best solution found so far ($\infty$ initially), and in first-solution mode the search stops when the first solution is recorded.

\textsc{Insert}$(Q', N, \tau)$ is \lacamstar{}'s successor handling. If $Q' \in$ EXPLORED, the edge $N \to Q'$ is added, costs are rewired, and the existing node is pushed again, or, with a small probability, the start node $N_s$ (\lacamstar{}'s random restart); the temporary goals $\tau$ are discarded. Otherwise a child $N'$ with $N'.Q = Q'$ is created and pushed, with $N'.\tau$ obtained from $\tau$ by ageing every record and applying the release rule (Eq.~\eqref{eq:release}).

\gb{} changes the loop in three places. When the unconstrained step at $N$ jams (Line~\ref{line:jam}), the blocking pairs $F$ of that step are turned into a queue of goal branches $N.\mathcal{B}$ (Sec.~\ref{sec:branches}). If that step reached an explored node other than $N$, \textsc{Insert} pushes that node, and $N$ is then pushed back on top of it (Line~\ref{line:push}). Whenever $N$ is at the top of OPEN and $N.\mathcal{B}$ is non-empty, one branch is tried before $N$'s next constraint (Lines~\ref{line:br-start}--\ref{line:br-end}): its temporary goals are written into a copy of $N.\tau$ with age $0$ and $d_i$ set to the pusher's current distance to its goal, its agents are placed first in the order, and PIBT is run without constraints.

The first branch of $N$ is therefore tried at the iteration after the jam: when the jam is a self-loop ($Q' = N.Q$, no agent moves), \textsc{Insert} pushes $N$ itself again, and otherwise $N$ is pushed back above the explored node. Later branches are consumed one per visit of $N$. A new successor is pushed on top of $N$, so the search continues from it, and the remaining branches are tried only if the search returns to $N$.

None of these changes affects how nodes are identified (by $Q$ alone), the structure of the constraint tree, or the costs, which is what the analysis in Sec.~\ref{sec:theory} relies on.

\subsection{Worked Example}\label{sec:example}
\begin{figure*}[t]
\centering
\includegraphics[width=\textwidth]{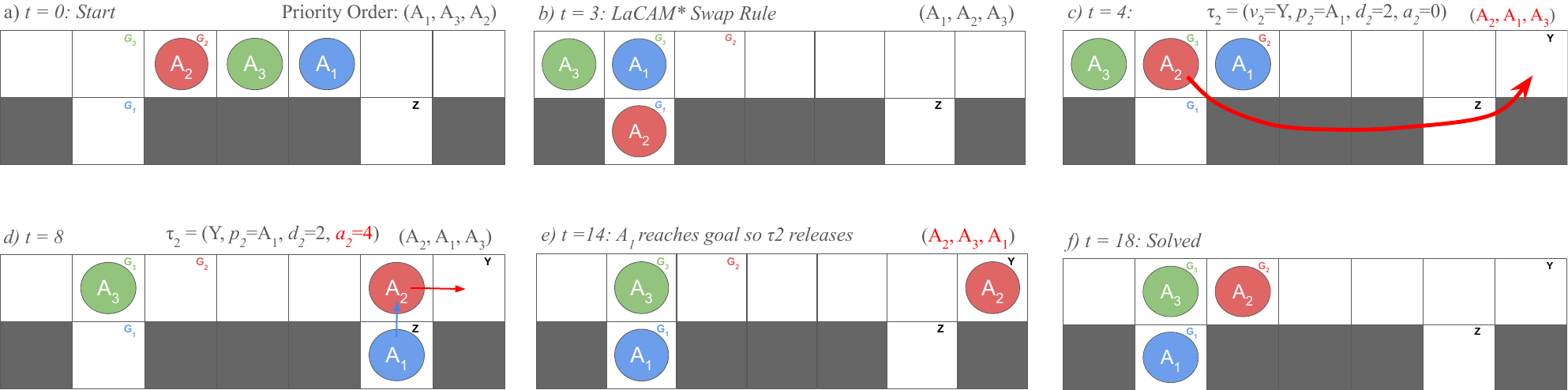}
\caption{Worked example (\gb{} with the random rule). Circles are agents and $G_i$ is the goal of agent $A_i$. (a)~Start. (b)~The swap technique pulls $A_2$ out of the pocket at $t{=}3$, but at $t{=}4$ the highest-priority agent $A_1$ pushes it back, so the step returns to an explored configuration. (c)~\gb{} gives $A_2$ the temporary goal $Y$ with pusher $A_1$. (d)~$A_2$ leads, $A_1$ steps into the pocket $Z$, and $A_2$ passes to $Y$. (e)~$A_1$ reaches its goal and $\tau_2$ is released. (f)~Final solved configuration.}
\label{fig:example}
\end{figure*}

Figure~\ref{fig:example} shows \gb{} on a seven-cell corridor with two pockets. $A_2$ starts on its goal at the mouth of a dead end that $A_1$ and $A_3$ must both enter, so $A_2$ has to leave the dead end entirely and get past $A_1$, which is possible only at the far pocket $Z$. \pibt{} instead pushes all three agents into the dead end in the wrong order, and from $t{=}3$ the search oscillates: the swap technique pulls $A_2$ out, and $A_1$, first in the priority order, pushes it straight back~(b). At this jam every push succeeds: $A_1$ pushes $A_2$, and $A_2$ can still move back into the pocket, so \pibt{} records no blocking pair. The fallback is used instead, and since $A_1$'s best next cell is occupied by $A_2$, it gives the pair $(A_2, A_1)$, and the first branch gives $A_2$ the temporary goal $Y$ with pusher $A_1$ and $d_2 = 2$~(c). As a holder, $A_2$ is now planned first. It drives $A_1$ back along the corridor, the two pass using the pocket $Z$, and $A_2$ waits at $Y$ while $A_1$ returns, since arriving at $Y$ does not release a temporary goal that has a pusher~(d). When $A_1$ reaches its goal at $t{=}14$, $\tau_2$ is released~(e), and $A_2$ returns to its own goal~(f). \gb{} solves this instance in 20 search iterations with a single branch, while \lacamstar{} needs between 1,477 and 3,387 iterations (depending on seed), since the detour takes $A_2$ four cells away from its goal and \pibt{} never proposes it.

This run is the best case for \gb{}. In about half of the runs that reach this jam, a random tie-break makes $A_2$ push $A_3$ first, and that failed push becomes the only blocking pair $(A_3, A_2)$, so every branch goes to $A_3$ and none resolves the jam. Even with $A_2$ first, only two of the six free vertices, $Y$ and $Z$, take it past $A_1$, so a random temporary goal works only a third of the time; over all runs, the first branch succeeds in about one in five. The outcome is still reliable, because a poor temporary goal at worst leads to another jam with a fresh set of branches, and the constraint tree remains as a fallback. Over 20 seeds, \gb{} needed at most 618 search iterations (median 66.5), against at least 1,477 for \lacamstar{}.

\subsection{Theoretical Properties}\label{sec:theory}
Let $\mathcal{Q}$ be the (finite) set of configurations. Each jam fills $N.\mathcal{B}$ at most once, with at most $B_{\max}$ branches
(Sec.~\ref{sec:branches}).

\begin{lemma}[\citealp{okumura2023lacam}]\label{lem:tree}
For any node $N$ and any fixed agent order $N.\pi$, if $N.\mathcal{T}$ is consumed entirely, every configuration $Q'$ with $N.Q \to Q'$ is generated as a successor of $N$.
\end{lemma}
\begin{proof}
A low-level node at depth $|A|$ constrains every agent, one per level in the order $N.\pi$, and every combination of moves appears at that depth. The constrained configuration is exactly $Q'$, and it is generated whenever it is collision-free.
\end{proof}

\begin{theorem}[Completeness]\label{thm:complete}
For any temporary-goal selection rule, \lacamstar{} with \gb{} returns a solution in finite time if one exists, and otherwise terminates reporting that none exists.
\end{theorem}
\begin{proof}
\emph{Termination.} Each iteration either pops a node from OPEN, consumes one branch from some $N.\mathcal{B}$, or consumes one low-level node from some $N.\mathcal{T}$, and pushes at most one node, plus $N$ itself after a jam (at most once per node), plus nodes whose $g$-value strictly decreases during rewiring, which happens finitely often since $g$-values are non-negative integers. $\mathcal{Q}$ is finite and each configuration is inserted into the explored set at most once, so there are finitely many high-level nodes. Each has a finite constraint tree and at most $B_{\max}$ branches, so the consumable items are finite, the number of pushes is finite, and the search terminates.

\emph{Correctness of a negative answer.} Suppose the search terminates with OPEN empty and no goal found. A node leaves OPEN only once $N.\mathcal{T}$ is exhausted. (Before a solution exists, no node is discarded by the $f$-bound.) Every high-level node therefore had its constraint tree consumed. By Lemma~\ref{lem:tree}, applied with each node's own fixed order, every successor of every generated configuration was itself generated. The generated configurations are thus closed under $\to$, and contain every configuration reachable from $Q_s$. Since $Q_g$ was not among them, no solution exists.

\emph{Validity of a positive answer.} Every edge in the returned path is either a constraint-tree successor or a branch successor. In both cases it is a configuration produced by PIBT from its parent, hence a collision-free transition. Temporary goals change which transitions PIBT proposes, never whether they are collision-free.

\end{proof}

\begin{proposition}[Eventual optimality]\label{prop:opt}
Under the conditions for which \lacamstar{} is eventually optimal~\citep{okumura2023lacamstar}, so is \lacamstar{} with \gb{}.
\end{proposition}
\begin{proof}
Edge costs and $h$-values are computed from the real goals $Q_g$ and never from $N.\tau$. They are therefore the same functions of configurations as in \lacamstar{}, and $h$ remains admissible. Every branch successor $N.Q \to Q'$ is a valid transition and is recorded as a neighbour edge, either at creation or by rewiring, so \gb{} adds only true edges of the configuration graph. Nodes are discarded only when $f \ge f(\text{goal})$, exactly as in \lacamstar{}, and every other node has its constraint tree exhausted. By Lemma~\ref{lem:tree}, every edge \lacamstar{} would discover is therefore discovered. Dijkstra rewiring over a superset of correctly weighted edges converges to the same optimal cost, so the argument of~\citet{okumura2023lacamstar} applies unchanged.
\end{proof}

\begin{proposition}[Conservative extension]\label{prop:cons}
Fix an instance and a random seed. If, in the run of \lacamstar{}, the unconstrained PIBT step of every expanded node produces a configuration not yet explored, then \lacamstar{} with \gb{} generates the same sequence of nodes and returns the same solution.
\end{proposition}
\begin{proof}
By induction over iterations, the two runs have identical OPEN, explored set, node priorities and orders, and random-number state. With no jam, no branch is queued, so $N.\tau_i = \bot$ for every node and agent, and every effective goal equals the real goal. PIBT's move ranking, swap checks and random tie-breaks are then identical. The priority boost adds nothing, so $N.\pi$ is identical. Jam detection only queries the explored set and draws no random numbers. Releasing temporary goals is a no-op. Hence each iteration produces the same successor in both runs, and since the hypothesis guarantees that no jam occurs in the \gb{} run either, the induction goes through.
\end{proof}
\section{Experiments}\label{sec:exp}
\subsection{Setup}  
We implemented \gb{} on top of the reference implementation of \lacamstar{} (lacam2) in C++, so that the only differences from \lacamstar{} are those described in Section~\ref{sec:gb}. All experiments were run on an AMD Ryzen~5  7600X under WSL2. Each run used a single thread pinned to its own physical core, with four runs in parallel, a time limit of 30\,s and a memory limit of 5\,GB. A run that exceeded either limit counts as a failure. 

We used six $32 \times 32$ maps from the MovingAI benchmark \citep{DBLP:conf/socs/SternSFK0WLA0KB19}. Three of them contain narrow corridors and dead ends (maze-32-32-2, maze-32-32-4 and room-32-32-4), and three are open (random-32-32-10, random-32-32-20 and empty-32-32). The benchmark's scenarios contain at most half as many agents as free cells, so we generated 25 denser scenarios per map, drawing starts and goals uniformly at random from the free cells and using the first $n$ agents of each scenario for an instance with $n$ agents. Densities range from 60\% to 90\% of the free cells on the bottleneck maps and from 70\% to 95\% on the open maps, in steps of 5\%. Each instance is run with five random seeds (0--4), giving 125 runs per map and density. Tables give values over all runs and, where useful, the standard deviation across the five seeds.

We compare \gb{} with the random and corridor rules ($K{=}3$, $P{=}4$, $H{=}64$) against \lacamstar{} (lacam2, commit 61a4c40) and LaCAM3 \citep{DBLP:conf/atal/000124} (commit 1c3ecf8), all stopping at their first solution. Both baselines use their default parameters, except that LaCAM3 is run single-threaded and without its anytime refinement (\texttt{--no-star --no-multi-thread}). We report the success rate within the time and memory limits, the time to the first solution, and the sum of costs divided by the sum of start-to-goal distances, a lower bound on the optimal sum of costs.

\subsection{Dense Instances}  
\begin{figure*}[t]
\centering
\includegraphics[width=\textwidth]{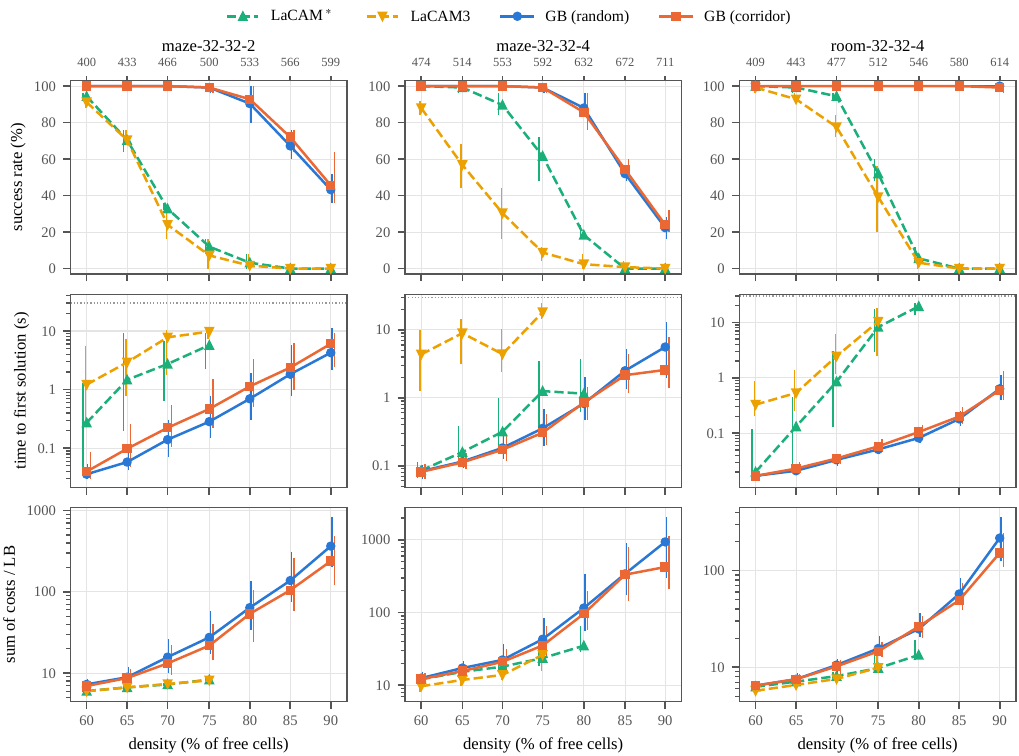}
\caption{Dense bottleneck maps: success rate (top), time to first solution (middle) and sum of costs relative to the lower bound (bottom) against density, with agent counts along the top. Time and cost are medians over solved runs, shown where at least 5 runs were solved; the dotted line is the 30\,s time limit. 125 runs per point (25 scenarios, five seeds). Error bars: range of the per-seed success rate (top) and interquartile range (middle, bottom).}
\label{fig:main}
\end{figure*}

Figure~\ref{fig:main} and Table~\ref{tab:main} show the results on the three bottleneck maps. \gb{} solves 89\% of the 2{,}625 runs with either goal rule, against 44\% for \lacamstar{} and 33\% for LaCAM3. The advantage holds on every map and grows with density. \lacamstar{} solves fewer than half of the runs from 70\% density on maze-32-32-2 and from 80\% on the other two maps, and none above 80\%, while \gb{} still solves 52--72\% of the runs at 85\% on the mazes and the random rule solves every run on room-32-32-4 at 90\%. \gb{} with the random rule solved 1{,}161 runs that \lacamstar{} did not, and \lacamstar{} solved none that it did not.

\begin{figure}[t]
\centering
\includegraphics[width=\columnwidth]{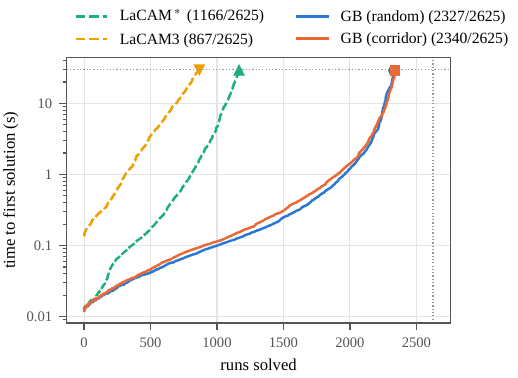}
\caption{Runs solved against time to first solution, pooled over the three bottleneck maps (2{,}625 runs per solver). A point at $(x, y)$ means the solver solved $x$ runs within $y$ seconds; the legend gives the number solved within 30\,s.}
\label{fig:cactus}
\end{figure}

Time and cost in Figure~\ref{fig:main} are taken over each solver's own solved runs, so once a solver starts failing they reflect only the instances it could still solve. Table~\ref{tab:main} therefore also compares \gb{} and \lacamstar{} on the runs both solved. On these runs \gb{} finds its first solution 13.8 times faster on maze-32-32-2 and 10.6 times faster on room-32-32-4 (geometric means), and at 75\% density on room-32-32-4 the median time falls from 8.2\,s to 0.04\,s. On maze-32-32-4 the gain on the runs both solve is smaller (1.9 times faster on average, with a median of 1.21), and most of the gain there is in the runs \lacamstar{} cannot solve at all. Figure~\ref{fig:cactus} shows the same picture as a function of the time budget. LaCAM3 is slower than \lacamstar{} throughout, because before searching it computes spread-out guide paths for the agents, which costs time on every instance.

Solutions found by \gb{} cost 22\% more than those of \lacamstar{} on the runs both solved (10\% on room-32-32-4 and 26--36\% on the mazes). On the densest instances, where \lacamstar{} solves fewer than five of the 125 runs, first solutions are much longer, with median costs of about 60 to 360 times the lower bound (random rule), and over 900 times at 90\% on maze-32-32-4. The two goal rules perform similarly. The random rule solves 2{,}327 runs and the corridor rule 2{,}340, with 79 and 92 runs solved by only one of them ($p = .36$), and the corridor rule's solutions are 11\% cheaper, at a cost of 18\% more time.

\begin{table}[t]
\centering
\small
\setlength{\tabcolsep}{2pt}
\begin{tabular}{lrrrr|rr|rr}
\hline
 & \multicolumn{4}{c|}{Solved (\%)} & \multicolumn{2}{c|}{Speed-up} & \multicolumn{2}{c}{Cost} \\
Map & L$^\ast$ & L3 & GB-r & GB-c & GB-r & GB-c & GB-r & GB-c \\
\hline
maze-2  & $30_{\pm 1}$ & $28_{\pm 1}$ & $86_{\pm 2}$ & $87_{\pm 2}$ & 13.8$\times$ & 9.7$\times$ & 1.36 & 1.30 \\
maze-4  & $53_{\pm 2}$ & $27_{\pm 3}$ & $80_{\pm 1}$ & $80_{\pm 1}$ & 1.9$\times$ & 2.0$\times$ & 1.26 & 1.14 \\
room-4  & $50_{\pm 1}$ & $45_{\pm 2}$ & $100_{\pm 0}$ & $100_{\pm 0}$ & 10.6$\times$ & 9.7$\times$ & 1.10 & 1.08 \\
\hline
all     & $44_{\pm 1}$ & $33_{\pm 1}$ & $89_{\pm 1}$ & $89_{\pm 1}$ & 5.7$\times$ & 5.2$\times$ & 1.22 & 1.15 \\
\hline
\end{tabular}

\caption{Dense bottleneck maps, 60--90\% density, 875 runs per map; maze-2, maze-4 and room-4 are maze-32-32-2, maze-32-32-4 and room-32-32-4. L$^\ast$/L3: \lacamstar{}/LaCAM3; GB-r/GB-c: \gb{} with the random/corridor rule. Speed-up and cost relative to \lacamstar{} (speed-up $>1$ is faster): geometric mean over runs that both \lacamstar{} and the GB variant solved. Subscripts: standard deviation of the solved \% across the five seeds. GB-r solved 1{,}161 runs that \lacamstar{} did not, and \lacamstar{} none that GB-r did not.}
\label{tab:main}
\end{table}

\subsection{Open Maps}  
On the open maps, all solvers solve 98--100\% of the runs, and \gb{} behaves almost exactly like \lacamstar{} (Table~\ref{tab:open}). On random-32-32-10 and empty-32-32 it takes at most 3\% more time on the runs both solve, and its solutions cost the same; 84--91\% of its plans are identical to those of \lacamstar{}, and, as Proposition~\ref{prop:cons} predicts, every run without a jam returns exactly \lacamstar{}'s plan. On random-32-32-20 jams are more common and only 28--29\% of plans are identical, but \gb{} is about 15\% faster there and its solutions are 1--4\% cheaper. LaCAM3 solves 99--100\% of the runs on these maps and finds cheaper first solutions, but takes 19 to 28 times longer than \lacamstar{}. 

\subsection{MovingAI Benchmarks}
Finally, we also ran the four solvers on all 33 maps of the MovingAI benchmark~\citep{DBLP:conf/socs/SternSFK0WLA0KB19}, with the 25 random scenarios of each map at 20, 40, 60, 80 and 100\% of the scenario's agents (at most 1,000), 4,125 runs per solver under the same limits with seed 0. All four solved every run on 32 of the maps. On the remaining map, maze-128-128-1, GB-r solved 121 of 125 runs against 80 for \lacamstar{}, 90 for GB-c and 69 for LaCAM3; GB-r solved 41 runs that \lacamstar{} did not, and \lacamstar{} none that GB-r did not. GB-r never jammed on 92.7\% of the runs it solved, and on every one of them it returned exactly the plan of \lacamstar{}. Over the 4,080 runs both solved, the geometric means of its time to first solution and cost relative to \lacamstar{} were 1.00 and 1.001. LaCAM3 was about ten times slower than \lacamstar{} on the runs both solved, but its plans were 19\% cheaper.
    
\begin{table}[t]
\centering
\small
\setlength{\tabcolsep}{3pt}
\begin{tabular}{lrrrr}
\hline
 & \multicolumn{2}{c}{Speed-up} & \multicolumn{2}{c}{Identical plans} \\
Map & GB-r & GB-c & GB-r & GB-c \\
\hline
random-32-32-10 & $0.99_{\pm 0.01}$ & $0.99_{\pm 0.01}$ & 84\% & 84\% \\
random-32-32-20 & $1.16_{\pm 0.04}$ & $1.15_{\pm 0.04}$ & 29\% & 28\% \\
empty-32-32     & $0.98_{\pm 0.01}$ & $0.98_{\pm 0.01}$ & 91\% & 91\% \\
\hline
\end{tabular}

\caption{Open maps, 70--95\% density, 750 runs per map. All solvers solve 98--100\% of runs. Speed-up: \lacamstar{}'s time to first solution over \gb{}'s, on runs both solved (geometric mean with SD across seeds; $>1$ is faster); cost is within 0.96--1.00. Identical plans: runs where \gb{} returns exactly \lacamstar{}'s plan.}
\label{tab:open}
\end{table}

\subsection{Ablation}\label{sec:ablation}
We evaluated variants of \gb{} on the three bottleneck maps at three high densities each (70--80\% on maze-32-32-2, 75--85\% on maze-32-32-4 and 80--90\% on room-32-32-4), with the same 25 scenarios and three random seeds each, giving 675 runs per variant (Table~\ref{tab:ablation}). Trying a jammed node's first branch immediately matters most. Without it, 28 fewer runs are solved ($p < .001$), with fewer on every map, and on the runs both solve, first solutions take about twice as long and cost 31\% more. The number of temporary goals per blocking agent ($K$) and the number of blocking agents per jam ($P$) make no significant difference, so \gb{} is not sensitive to how many branches it queues, and shortening the hold limit to $H=16$ solves exactly as many runs. Restricting random goals to within distance 8 of the blocking agent, however, solves 131 fewer runs, which suggests that good temporary goals are often far from the blocking agent. We also tested two other ways of choosing blocking pairs. A failure recorded during \pibt{} is \emph{absorbed} if an agent further up the same chain still obtains a vertex as good as its best option, as when a pushed agent fails but its pusher simply takes another, equally good vertex. Such failures may suggest temporary goals for agents that were not really in the way, so the first variant discards them (no absorbed). The second ignores \pibt{}'s failures altogether and always uses the fallback pairs of Section~\ref{sec:jam} (fallback only). Discarding absorbed failures makes no difference (one more run solved, $p = 1.0$). Using only the fallback pairs solves 16 fewer runs ($p = .023$; more on maze-32-32-2 but fewer on the other maps) and roughly doubles the cost of solutions, so the pairs taken from \pibt{}'s failures lead to cheaper solutions and, overall, to more solved runs. The corridor rule solves about as many runs as the random rule (one more, $p = 1.0$), and its solutions are 16\% cheaper, at a cost of 22\% more time. Removing recursion from the corridor rule makes no difference (two more runs solved, $p = .69$), but then removing retreat as well solves 31 fewer runs ($p < .001$), and its solutions cost 21\% more.

\begin{table}[t]
\centering
\small
\setlength{\tabcolsep}{3pt}
\begin{tabular}{lrrrrr}
\hline
Variant & Solved & $\Delta$ & $p$ & Speed-up & Cost \\
\hline
\gb{} (default)                    & 622 & --     & --      & 1.00 & 1.00 \\
No immediate branches              & 594 & $-28$  & $<$.001 & 0.51 & 1.31 \\
$K = 1$                            & 619 & $-3$   & .70     & 0.97 & 1.02 \\
$K = 5$                            & 614 & $-8$   & .22     & 1.00 & 1.01 \\
$P = 1$                            & 619 & $-3$   & .74     & 0.98 & 1.01 \\
$P = 8$                            & 616 & $-6$   & .39     & 1.00 & 1.03 \\
$H = 16$                           & 622 & $0$    & 1.0     & 0.99 & 0.97 \\
Radius $R = 8$                     & 491 & $-131$ & $<$.001 & 0.48 & 1.08 \\
Pairs: no absorbed                 & 623 & $+1$   & 1.0     & 0.85 & 1.04 \\
Pairs: fallback only               & 606 & $-16$  & .023    & 0.68 & 1.94 \\
\hline
Corridor rule                      & 623 & $+1$   & 1.0     & 0.82 & 0.84 \\
\quad no recursion                 & 625 & $+2$   & .69     & 1.02 & 1.01 \\
\qquad no retreat                  & 594 & $-31$  & $<$.001 & 0.87 & 1.21 \\
\hline
\end{tabular}

\caption{Ablation on 675 dense runs (maze-32-32-2, maze-32-32-4, room-32-32-4; three densities each). Default: random rule, $K{=}3$, $P{=}4$, $H{=}64$. No immediate branches: a jammed node's goal branches are tried only when the search returns to it. $\Delta$: solved relative to the default, or for the indented rows, relative to the row above; $p$: McNemar exact test; Speed-up: time to first solution of the reference over the variant's ($>1$ is faster); Cost: sum of costs relative to the reference. Speed-up and cost are geometric means over runs both solved.}
\label{tab:ablation}
\end{table}

\section{Discussion}

Overall, \gb{} consistently improves \lacamstar{} on dense instances with corridors and dead ends, both in the number of instances solved and in the time to a first solution, and costs almost nothing on open maps, where it is never more than 3\% slower and is faster on random-32-32-20. The random rule is as reliable as the corridor rule on the dense maps and more reliable on maze-128-128-1, and the method is insensitive to how many branches it creates. This suggests that \gb{}'s strength lies less in choosing a good temporary goal than in the search's ability to try several as soon as a jam occurs and recover from poor ones, as long as temporary goals can be far from the blocking agent. The corridor rule finds cheaper solutions but is slower, and its pocket search cannot look past the pusher, which limits it on instances like our worked example. The main cost of \gb{} is solution quality, since its first solutions are longer than those of \lacamstar{}, although \lacamstar{}'s anytime refinement applies unchanged. Considering these observations, our suggestion is to use the random rule with its default parameters, and the corridor rule when solution cost matters more than runtime.

Implementing \gb{} on top of \lacamstar{} is the safe choice because the existing constraint framework means that temporary goals that lead to failure or to repeated configurations are handled by the high-level search. We are also interested in applying this approach to \pibt{} alone, without \lacamstar{}'s guarantees, to understand whether it can reduce or even eliminate deadlocks in \pibt{}. This will require a much more careful selection of temporary goal locations, since there is no search to recover from a poor choice.

Temporary goal branches can be thought of as a single example of inserting informed branches into \lacamstar{}'s search ahead of its constraint tree. Our completeness proof only requires that each node receives finitely many extra successors and that each of them is a valid transition, so other kinds of branches could be added in the same way. Other examples could include evacuation plans from rule-based planners such as MACGA, rotations of several agents around a cycle, or \pibt{} steps that use a different tie-breaking rule such as hindrance.

We also believe that intelligent heuristics for the location of temporary goals could improve \gb{} further. Our corridor rule does not solve significantly more instances than random goals, although its solutions are 11--16\% cheaper, and restricting random goals to nearby vertices makes \gb{} much worse (Section~\ref{sec:ablation}), which suggests that good temporary goals are often far from the blocking agent. The worked example shows why, because the corridor rule only looks for pockets on the near side of the pusher and so can never send $A_2$ past $A_1$ to the far pocket. A rule that can look past the pusher, or one that learns where temporary goals tend to succeed, might keep the reliability of random goals while lowering solution cost.

\section{Conclusion}
We presented \gb{}, an extension of \lacamstar{} that responds to jams in \pibt{} by giving blocking agents temporary goals and trying each choice as a separate branch of the search. \gb{} preserves completeness and eventual optimality for any rule for choosing temporary goals. On dense instances with corridors and dead ends it solves substantially more instances than \lacamstar{}, often an order of magnitude faster, and on the full MovingAI benchmark it never solves fewer runs and matches \lacamstar{}'s cost and speed on average. Promising directions for future work are goal-selection rules that reduce solution cost, and combining \gb{} with methods that improve \pibt{}'s move preferences, such as hindrance.

\bibliography{aaai2027}


\end{document}